\documentclass[twoside,11pt]{article}

\usepackage{jmlr2e}

\usepackage{amsmath, amsfonts}  
\usepackage{booktabs}        
\usepackage{algorithm}
\usepackage{algorithmic}
\usepackage{xfrac}              
\usepackage{enumerate} 

\usepackage{amsmath}      

\usepackage{bm}

\usepackage{color}
\definecolor{Orange}{rgb}{1,0.5,0}

\usepackage{array}
\newcolumntype{H}{>{\setbox0=\hbox\bgroup}c<{\egroup}@{}}
\usepackage{multicol}

\usepackage[T1]{fontenc}

\newcommand{\I}{\mathsf{I}}

\firstpageno{1}

\usepackage{lastpage}
\jmlrheading{21}{2020}{1-\pageref{LastPage}}{4/19; Revised
9/20}{10/20}{19-307}{Reza Mohammadi, Matthew Pratola and Maurits Kaptein}
\ShortHeadings{Continuous-Time Birth-Death MCMC for Bayesian Regression Tree Models}{Mohammadi, Pratola and Kaptein}

\begin{document}

\title{Continuous-Time Birth-Death MCMC for Bayesian Regression Tree Models}
\ShortHeadings{Continuous-Time Birth-Death MCMC for Bayesian Regression Tree Models}{Mohammadi, Pratola and Kaptein}

\author{
	  \name Reza Mohammadi \email A.Mohammadi@uva.nl \\
       \addr Amsterdam Business School \\
       University of Amsterdam \\
       Amsterdam, The Netherlands
       \AND
       \name Matthew Pratola \email mpratola@stat.osu.edu \\
       \addr Department of Statistics\\
       The Ohio State University\\
       Ohio, USA
       \AND
       \name Maurits Kaptein \email m.c.kaptein@uvt.nl \\
       \addr Statistics and Research Methods\\
       University of Tilburg\\
       Tilburg, The Netherlands}

\editor{Sayan Mukherjee}

\maketitle

\begin{abstract}
Decision trees are flexible models that are well suited for many statistical regression problems. In the Bayesian framework for regression trees, Markov Chain Monte Carlo (MCMC) search algorithms are required to generate samples of tree models according to their posterior probabilities. The critical component of such MCMC algorithms is to construct ``good'' Metropolis-Hastings steps to update the tree topology. Such algorithms frequently suffer from poor mixing and local mode stickiness; therefore, the algorithms are slow to converge. Hitherto, authors have primarily used \emph{discrete}-time birth/death mechanisms for Bayesian (sums of) regression tree models to explore the tree-model space. These algorithms are efficient, in terms of computation and convergence, only if the rejection rate is low which is not always the case. We overcome this issue by developing a novel search algorithm which is based on a \emph{continuous}-time birth-death Markov process. The search algorithm explores the tree-model space by jumping between parameter spaces corresponding to different tree structures. The jumps occur in continuous time corresponding to the birth-death events which are modeled as independent Poisson processes. In the proposed algorithm, the moves between models are \emph{always} accepted which can dramatically improve the convergence and mixing properties of the search algorithm. We provide theoretical support of the algorithm for Bayesian regression tree models and demonstrate its performance in a simulated example. 
\end{abstract}


\section{Introduction}

Classification and regression trees \citep{breiman1984classification} provide a flexible modeling approach using a binary decision tree via splitting rules based on a set of predictor variables. Tree models often perform well on benchmark data sets, and they are, at least conceptually, easy to understand \citep{de2000classification}. Tree-based models, and their extensions such as ensembles of trees \citep{prasad2006newer} and sums of trees \citep{chipman2010bart} are an active research area and arguably some of the most popular machine learning tools\citep{biau2012analysis, biau2008consistency, chipman1998bayesian, denison1998bayesian, chipman2002bayesian, wu2007bayesian, linero2018bayesian, au2018random, probst2017tune, pratola2014parallel}. 

Much contemporary research work has focused on Bayesian formulations of regression trees \citep[see, e.g.,][]{denison1998bayesian,chipman2010bart}. The Bayesian paradigm provides, next to a good predictive performance, a principled method for quantifying uncertainty \citep{robert2007bayesian}. This Bayesian formulation can, amongst other uses, be extremely valuable in sequential decision problems \citep{robbins1985some, gittins2011multi} and active learning \citep{cohn1996active} for which popular approaches include Thompson sampling \citep{thompson1933likelihood, agrawal2012analysis}. It is vital to know not merely the expected values (or some other point estimate) of the modeled outcome, but rather to obtain a quantitative formulation of the associated uncertainty \citep{eckles2014thompson, eckles2019bootstrap}. This is exactly what Bayesian methods readily provide \citep{robert2007bayesian}.

Recent Bayesian formulations of regression trees have already found their way into many applications \citep{gramacy2008bayesian}, but computationally efficient sampling algorithms for tree models and sum-of-tree models have proven non-trivial: the tree-model space of possible trees grows rapidly as a function of the number of features and efficient exploration of this space has proven cumbersome \citep{pratola2016efficient}. Numerous methods have been proposed to address this problem; indeed, the popular sums-of-trees model specification proposed by \citet{chipman2010bart} is itself an attempt to reduce the tree depth and thereby partly mitigate the problem. Other recent approaches have focussed on efficiently generating Metropolis-Hasting (MH) proposals in the Markov Chain Monte Carlo (MCMC) algorithm \citep[see][for examples]{pratola2016efficient, wu2007bayesian}, or alternatives to the MH sampler such as sequential MCMC \citep{taddy2011dynamic} and particle-based approaches \citep{lakshminarayanan2013top}.

To the best of our knowledge, the most effective search algorithm known at this point in time is provided by \citet{pratola2016efficient}, who efficiently integrates earlier advances and adds a number of novel methods to generate tree proposals. \citet{pratola2016efficient} implements these methods to explore the tree-space by using a search algorithm that is known as reversible jump MCMC (RJ-MCMC) \citep{green1995reversible} which is based on an ergodic, \emph{discrete}-time Markov chain. The RJ-MCMC algorithm often suffers from high rejection rates especially when the model space is large which is the case for the decision tree models. Therefore, these algorithms often are poor mixing and slow to converge.

In this paper, to overcome this issue, we make a significant contribution to the Bayesian decision tree literature by proposing a novel \emph{continuous}-time MCMC (CT-MCMC) search algorithm which is essentially the continuous version of the RJ-MCMC algorithm. The main advantage of the CT-MCMC algorithm is that each step of the MCMC algorithm considers the whole set of transitions and a transition always occurs; in fact, there is no rejection. Thus, the CT-MCMC has clearly better performances in terms of computational time and convergence rate. The proposed CT-MCMC search algorithm is based on the construction of continuous-time Markov birth-death processes (introduced by \citealt{preston1976}) with the appropriate stationary distribution. Sampling algorithms based on these processes have already been used successfully in the context of mixture distributions by \cite{stephens2000bayesian, cappe2003reversible, mohammadi2013using}. In the case of mixture distributions, the birth-death mechanisms have been implemented in the MCMC algorithm in such a way that the algorithm explores the model space by adding/removing a component for the case of a birth/death event. More recently, such MCMC algorithms have been used in the field of (Gaussian) graphical models \citep{mohammadi2015bayesian, mohammadi2017bayesian, dobra2018loglinear, wang2020scalable, dobra2018loglinear, hinne2014efficient, mohammadi2019bdgraph, mohammadi2017ratio}. For the case of graphical models, the birth-death mechanisms have been implemented in the MCMC algorithm in such a way that the algorithm explores the graph space by adding/removing a link for the case of a birth/death event. 

We apply this continuous-time MCMC mechanism to the classification and regression tree (CART) model context, by considering the parameters of the model as a point process, in which the points represent the nodes in the tree model. The MCMC algorithm explores the tree space by allowing new terminal nodes to be \textit{born} and existing terminal nodes to \textit{die}.  These birth and death events occur in continuous time, as independent Poisson processes; see Figure~\ref{fig:CT-MCMC}. We design the MCMC algorithm in such a way that the relative rates of the birth/death events determine the stationary distribution of the process. In Section \ref{sec:ct-mcmc} we formalize the relationship between the birth/death rates and the stationary distribution. Based on this we construct the MCMC search algorithm in which the birth/death rates are the ratios of the posterior distributions. We show how to use the advantage of continuous-time sampling to efficiently estimate the parameter of interest based on model averaging, using Rao-Blackwellization \cite{cappe2003reversible}. 

This paper is structured as follows. In the next section we introduce the tree and sums of tree models more formally and introduce the sampling challenges associated with this model in more detail. Next, in Section \ref{sec:ct-mcmc} we detail our suggested alternative birth-death approach and provide both an efficient algorithm and the theoretical justification for our proposal. Subsequently, we extend this proposal to also include the rotation moves suggested by \citet{pratola2016efficient}. In Section \ref{sec:evaluation} we compare the performance of our method---in terms of both its statistical properties and its computation time---to the current state of the art \citep{pratola2016efficient} using a simple, well-known, example that is notoriously challenging for tree models \citep{wu2007bayesian}. Finally, in Section \ref{sec:discussion} we discuss the limitations of our contribution and provide pointers for future work.

\section{Bayesian tree models}
\label{sec:notation}

\begin{figure}[ht]
\centering
\includegraphics[width=0.6\textwidth]{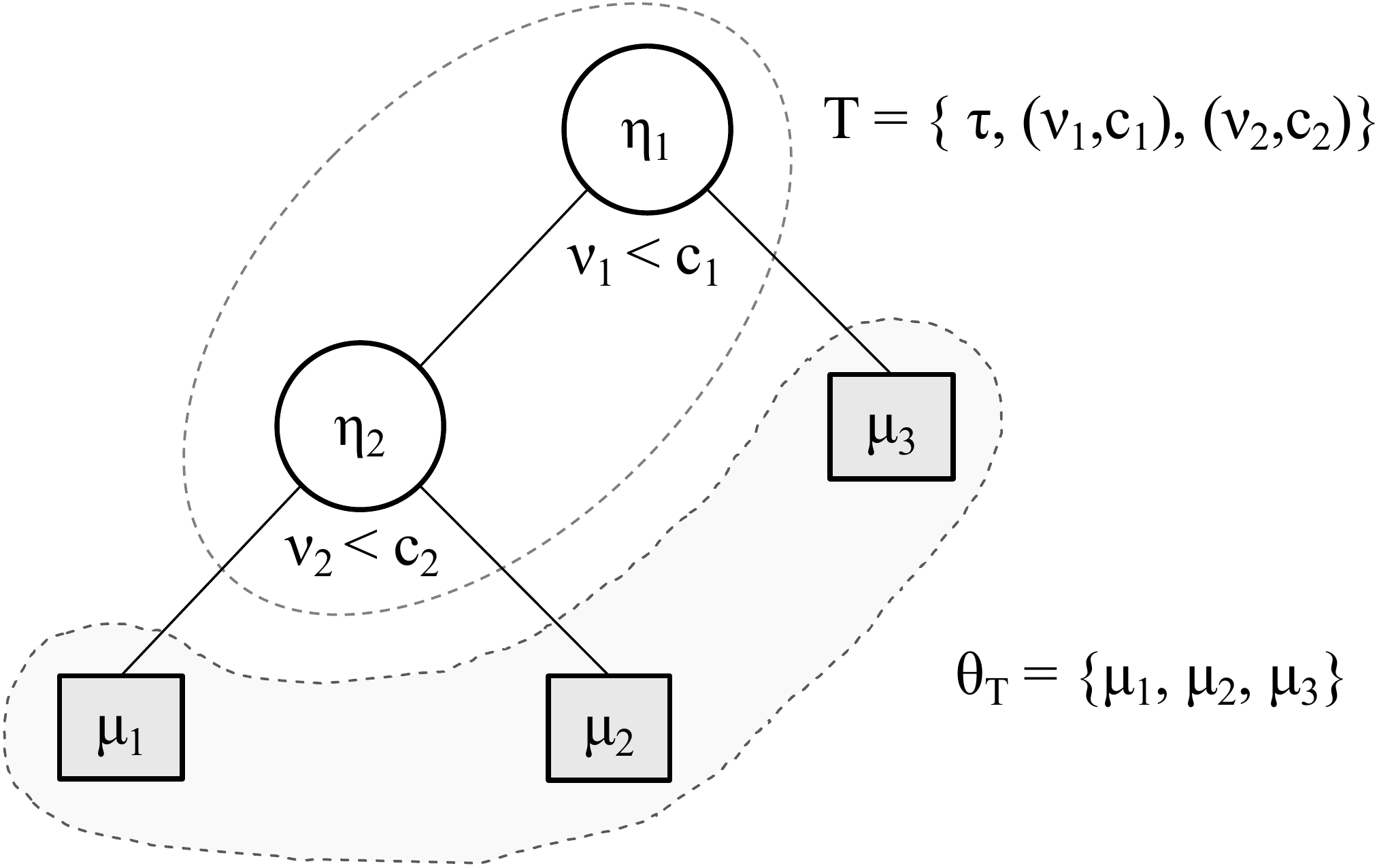}
\caption{A simple example of our main notation for a tree model which has $2$ interior nodes ($\eta_1, \eta_2$). $T$ encodes both the split rules $\{ (\nu_1, c_1), (\nu_2, c_2), \dots \}$ as well as the topology $\tau$. The set of maps $\theta_T = \{\mu_1, \mu_2, \dots, \mu_{n_t} \}$ determines the values of the terminal nodes. 
}
 \label{fig:fig1}
\end{figure}

We consider binary regression or classification trees and sum of trees models. Given a feature vector $x=(x_1, ..., x_d)$, and a scalar output of interest $y$ we can denote the tree model as follows
\begin{eqnarray*}
y = g(x; T, \theta_T) + \epsilon,  \;  \; \; \; \; \; \; \; \; \; \; \epsilon \sim \mathcal{N}(0,\sigma^2)
\end{eqnarray*}
where $T$ denotes the interior nodes of the tree, $\theta_T$ denotes a set of maps associated with the terminal nodes. Effectively, $T$ encodes all the (binary) split rules that jointly generate the tree structure. This is often expressed using a list of tuples $\{ (\nu_1, c_1), (\nu_2, c_2), \dots \}$ where $\nu_i \in \{1, \dots, d \}$ indicates which element of the feature vector to split on, and $c_i$ denotes the associated value of the split \citep[see, e.g.,][]{pratola2016efficient}. This way of expressing the tree is however limited since it does not encode the actual topology $\tau$ of the tree, which encodes the number of nodes in a tree, whether a node is internal or terminal, parent/child edges, and node depths. Hence, more precisely $\tau$ and $\{ (\nu_1, c_1), (\nu_2, c_2), \dots \}$ \emph{jointly} make up the full tree structure $T$. Figure~\ref{fig:fig1} illustrates our notation at this point in the paper; in Section \ref{sec:ct-mcmc} we will gradually introduce some additional notation necessary for our theoretical justification.

Given the number of terminal nodes, $n_t$, the maps $\theta_T = \{\mu_1, \mu_2, \dots, \mu_{n_t} \}$ take as input a feature vector $x$ and produce a response $\mu_j(x)$. In typical tree regression models the maps are constants; $\mu_j(x) = \mu_j$. Taken together, $T$ represents a partitioning of the feature space and a mapping from an input feature $x$ to a response value encoded in $\theta_T$. 

The Bayesian formulation of the tree model is completed by using priors of the form 
\[
\pi(T, \theta_T, \sigma^2) = \pi(\theta_T | T)\pi(T)\pi(\sigma^2).
\]

Note that the sum-of-trees model \citep{chipman2010bart} provides a conceptually straightforward extension of the above specified single tree model
\begin{eqnarray}
\label{eq:sums}
y = \sum_{m=1}^{M} g(x; T_m, \theta_{T_m}) + \epsilon
\end{eqnarray}
where the sum runs over $M$ distinct trees whose outputs are added. In the case of the sum-of-trees model we have
\[
\pi( T_1, \theta_{T_1}, ... , T_M, \theta_{T_M}, \sigma^2) = \left[ \prod_{m=1}^M \pi(\theta_{T_m} | T_m)\pi(T_m) \right] \pi(\sigma^2).
\]
For more details related to the sum-of-tree models we refer to \cite{pratola2016efficient}.

\subsection{Specification of the tree prior}
\label{sec:prior}

We specify the prior $\pi(T)$ by three parts,

\begin{enumerate} 
\item[$-$] The distribution on the splitting variable assignments at each interior node $\nu$ as a discrete random quantity in $\left\{ 1, ..., d \right\}.$
\item[$-$] The distribution on the cut-point $c$  as a discrete random quantity in  
\[
\left\{ 0, \frac{1}{n_\nu - 1}, ..., \frac{n_\nu - 2}{n_\nu - 1} \right\} 
\]
where $n_\nu$ is the resolution of discretization for variable $\eta$.
\item[$-$] The prior probability that a node $i$ ($\eta_i$) at depth $d_i$ is non-terminal to be 
\[
\pi( \eta_i ) \propto \frac{\alpha}{(1+d_i)^\beta}, \;  \; \; \; \; \; \; \; \; \; \; \alpha \in (0, 1), \beta \geqslant 0.
\] 
\end{enumerate}

To specify the prior distributions on bottom-node $\nu$'s, we use standard conjugate form
\[
\theta_T=\{\mu_1, \mu_2, \dots, \mu_{n_t} \}| T \mathbin{\overset{iid}{\kern3pt \sim \kern3pt}} \mathcal{N}( \hat{\mu} , \sigma_{\mu}^2 ).
\] 
In practice, the observed $Y$ can be used to guide the choice of the prior parameter values for $(\hat{\mu}, \sigma_{\mu}^2)$; See e.g. \cite{chipman1998bayesian}. Here for simplicity we assume $\hat{\mu} = 0$. 

For a prior specification of the $\sigma$, we also use a conjugate inverse chi-square distribution prior
\[
\sigma^2 \sim \mathcal{IG} \left( \frac{\nu}{2}, \frac{\nu\lambda}{2} \right)
\] 
which results in simple Gibbs updates for the variance. In practice, the observed $Y$ can be used to guide the choice of the prior parameter values for $(\nu, \lambda)$; See e.g. \cite{chipman1998bayesian}.

\subsection{Sampling from tree-models}
\label{sec:sampling}

For a single tree the full posterior of the model, for given tree $(T, \theta_T)$, $\sigma$, and data $\mathcal{D}$ is 
\begin{equation}
 \label{eq:full posterior}
\Pr(T,\theta_{T}, \sigma \mid \mathcal{D}) \propto L( T, \theta_T, \sigma )  \pi(\theta_T \mid T) \pi(T)  \pi(\sigma).
\end{equation}

For a single tree---which could easily be extended to the sum-of-tree case---sampling from the full posterior of the model \ref{eq:full posterior} is conceptually carried out by iterating the following steps

\begin{enumerate}
\item Draw a new topology $\tau | y, \sigma^2,  \{(\nu_i, c_i)\}$ using some method of generating new topologies such as a birth/death or rotation and subsequently accepting or rejecting the proposal.
\item Draw the split rules $(\nu_i, c_i) | y, \tau, \sigma^2,\{ (\nu_{-i}, c_{-i}) \}$, $\forall i$ using perturb or perturb within change-of-variable proposals.
\item Draw $\mu_j | y, \tau, \sigma^2,\{ (\nu_{i}, c_{i}) \}$ using conjugate Gibbs sampling. 
\item Draw $\sigma^2 | y, \tau, \{\mu_j\},\{ (\nu_{i}, c_{i}) \}$ also using conjugate Gibbs scheme. 
\end{enumerate}

The above algorithm has been implemented successfully in earlier work \citep[see][]{pratola2016efficient}. Steps 3 and 4 are the standard Gibbs sampling using conjugate priors. Also Step 2 is efficiently implemented by \citep[Section 4]{pratola2016efficient}. For the sampling of $\tau | y, \sigma^2,  \{(\nu_i, c_i)\}$ (i.e., in step 1 above) the current state-of-the art is to use an RJ-MCMC search algorithm. In practice the RJ-MCMC algorithm performs well if the rejection rate is low \citep[the computation of which is detailed in Equation 4 of ][]{pratola2016efficient}. However, when the rejection rate is not low---which is often the case---the mixing of the chain is poor and the exploration of the full tree-model space is notoriously slow. To overcome this issue, in the next section, we introduce a novel search algorithm which has basically no rejection rate. So the novelty of our work mainly lies in the new search algorithm for step 1.

\section{Continuous-time birth-death MCMC search algorithm}
\label{sec:ct-mcmc}

\begin{figure}[htp]
\centering
\includegraphics[width=0.95\textwidth]{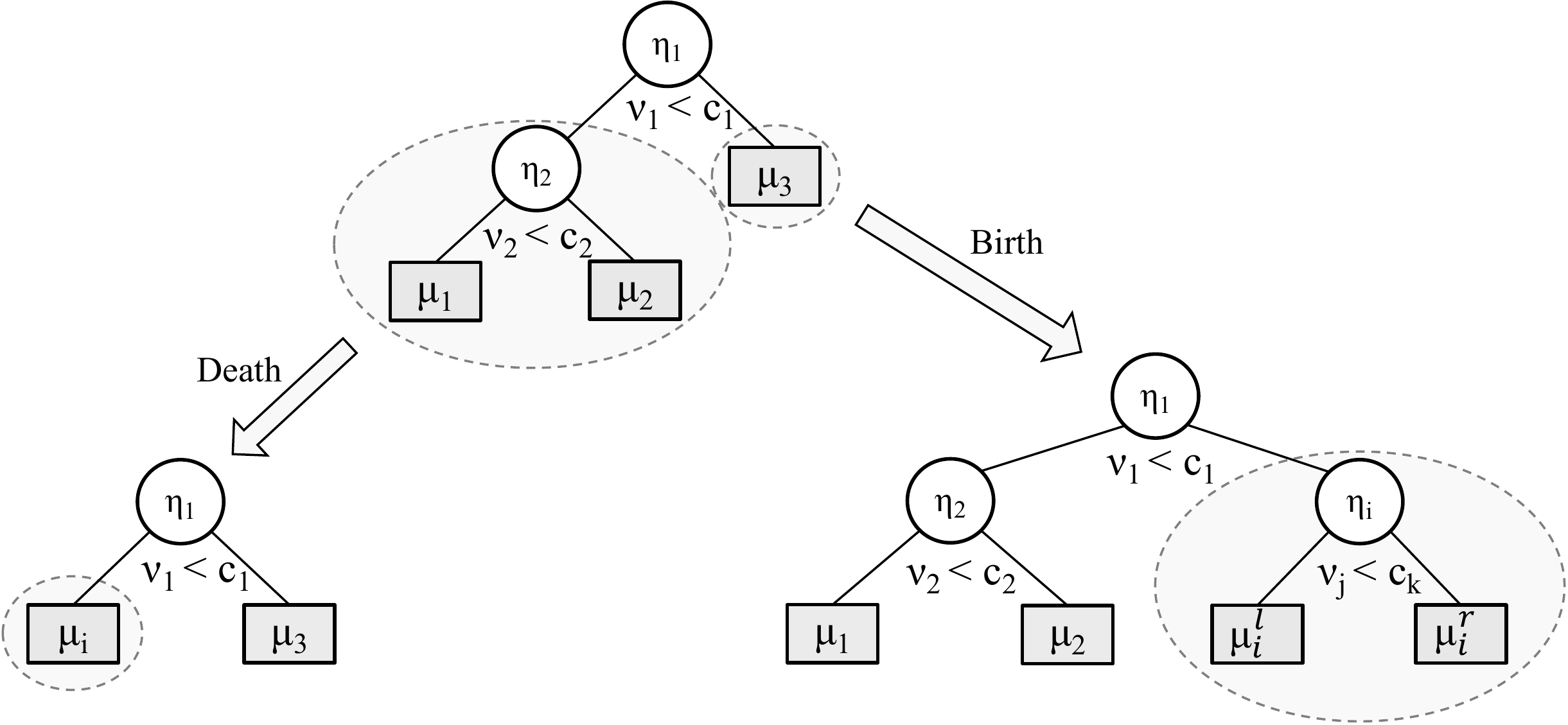}
\caption{The birth-death mechanism for adding or deleting nodes of the tree. On the bottom left a death occurs at node $\eta_2$ from the original resulting in the removal of maps $(\mu_2, \mu_3)$ and the emergence of a new map $\mu_i$. On the bottom right a new node $\eta_i$ is born at map $\mu_1$ resulting in the removal of this map and the addition of $\nu_j$, $c_k$, and $(\mu^l_i, \mu^r_i)$.}
 \label{fig:birth-death}
\end{figure}

The issue of a low acceptance rate in step 1 of the algorithm mentioned in the previous section is surprisingly common: as the tree space is extremely large, proposals with a low likelihood are frequent. This specific issue can however be overcome by adopting a \emph{continuous}-time Markov process---or a CT-MCMC search algorithm---as an alternative to RJ-MCMC. In this sampling scheme the algorithm explores the tree-model space by either jumping to a larger dimension (birth) or lower dimension (death) as in step 1 above. But this time each of these events is modeled as an independent Poisson process and the time between two successes events is exponentially distributed. The change events thus occur in continuous time and their rates determine the stationary distribution of the process; see Figure~\ref{fig:birth-death} for a graphical overview of possible birth and deaths from a given tree. Unlike the RJ-MCMC, in the CT-MCMC search algorithm the moves between models are \emph{always} accepted making the algorithm more efficient. 

\citet{cappe2003reversible} have shown, on appropriate re-scaling of time, that the RJ-MCMC converges to a continuous-time birth-death chain. One advantage of CT-MCMC is its ability to transit to low probability regions that can form a kind of ``springboard'' for the algorithm to flexibly move from one mode to another. 

Our strategy is to view each component of the terminal nodes of the tree as a point in parameter space, and construct a Markov chain with the posterior distribution of the parameters as its stationary distribution. For given tree $(T, \theta_T)$ and data $\mathcal{D}$, the target posterior distribution is 
\begin{equation}
 \label{eq:target posterior}
\Pr(T,\theta_{T}\mid \mathcal{D}) \propto L( T, \theta_T ) \pi(T) \pi(\theta_T)
\end{equation}
where $L(T, \theta_T)$ is the likelihood. Note that the proposed search algorithm for sampling the tree model can then be combined with conjugate Gibbs updates of the continuous parameters such as $\sigma^2$ similar to the Metropolis-within-Gibbs algorithm; see for example \cite{stephens2000bayesian}.

We take advantage of the theory on general classes of Markov birth-death processes from \citet*[Section 7 and 8]{preston1976}. This class of Markov jump processes evolve in jumps which occur a finite number of times in any finite time interval. These jumps are of two types: (i) \textit{birth} in which a single point is added, and the process jumps to a state that contains the additional point; and (ii) \textit{death} in which one of the points in the current state is deleted, and the process jumps to a state with one less point. \citet{preston1976} shows that this process converges to a unique stationary distribution provided if the detailed balance conditions hold. 

To properly define the birth and death events in our case we need to introduce some additional notation identifying the different nodes in the tree and their respective variables and cut-points. Let $(T, \theta_T)$ define the tree model as before, additionally let $n_t$ be the number of terminal nodes, $n_{\nu}$ the number of variables, and $n_c$ the number of cut-points. Given the current state $(T, \theta_T)$:
\begin{itemize}
\item[\textbf{Birth:}] A new terminal node is created (born) in continuous time with a \emph{birth rate} $B_{ijk}(T, \theta_T)$; we denote this operation by `$\cup$'. In this case, the process transits to a new state 
\[
(T^{b_{ijk}}, \theta_{T^{b_{ijk}}} )  = (T \cup (\eta_{i},\nu_{ij},c_{ijk}),   \theta_T \cup (\mu^l_i,\mu^r_i) \setminus \mu_i)
\]
where $\eta_{i}$ denotes to internal node $i$, $i \in 1, ..., n_t$, $j \in 1, ..., n_{\nu}$, and $k \in 1, ..., n_c$. Further, we define the total birth rate as 
\[
B(T, \theta_T) = \sum_{i=1}^{n_t} \sum_{j=1}^{n_{\nu}} \sum_{k=1}^{n_c} B_{ijk}(T, \theta_T). 
\]
Hence, a birth event changes the topology $\tau$ of the current tree $T$ by adding a terminal node $i$. Accordingly, to complete the specification of the new tree $(T^{b_{ijk}}, \theta_{T^{b_{ijk}}})$ we also need to add variable $\nu_j$ and cut-point $c_k$ as well as the new terminal maps $(\mu^l_i,\mu^r_i)$. This process is illustrated in Figure~\ref{fig:birth-death} on the bottom right where a birth occurs at map $\mu_1$.  

\item[\textbf{Death:}] In the current state $(T, \theta_T)$ with $n_d$ terminal nodes, one of the terminal nodes is killed in continuous time with \emph{death rate} $D_{i}(T, \theta_T)$; we denote this operation by `$\setminus$' . In this case, the process transits to state 
\[
(T^{d_{i}},  \theta_{T^{d_{i}}}) = (T \setminus (\eta_{i}, \nu_{i},c_{i}),\theta_T\setminus ( \mu_{i}^{l},\mu_{i}^{r} ) \cup\mu_i)
\]
where $i \in 1, ..., n_d$ and $n_d$ is the number of possible deaths. Also, we define the total death rate as
\[
D(T, \theta_T) =  \sum_{i=1}^{n_d} D_{i}(T, \theta_T).
\]
Hence, a death event changes the topology $\tau$ by removing node $i$, including its associated variable and cut-point, $(\nu_i, c_i)$ and their respective maps $( \mu_{i}^{l},\mu_{i}^{r} )$. Accordingly, to complete the specification of the tree, we need to add a new map $\mu_i$. This process is illustrated in Figure~\ref{fig:birth-death} on the bottom left where a death occurs at node $\eta_2$.  
\end{itemize}

Since birth and death events are independent Poisson processes, the time between two consecutive events has an exponential distribution with mean
\begin{equation}
 \label{waiting time}
W(T, \theta_T) = \frac{1}{B(T, \theta_T)+D(T, \theta_T)}
\end{equation} 
which is the \textit{waiting time}. Note that the waiting times are calculated based on all the possible birth and death moves from the current state $(T, \theta_T)$ to a new state which would be a tree with one more or less terminal nodes regarding to the birth/death rates. Therefore, the waiting times essentially capture all the possible moves of each step of the CT-MCMC search algorithm. If the waiting time from $(T, \theta_T)$ is large then the process tends to stay longer in the current state while if the waiting time is small, the process tends to transition away from the current state. The birth and death probabilities involved are 
\begin{equation}
 \label{prob.birth}
 \Pr( \mbox{birth at node $\eta_i$ for variable $\nu_j$ and cut-point $c_k$} ) = \frac{B_{ijk}(T, \theta_T)}{ B(T, \theta_T) + D(T, \theta_T)}, 
\end{equation}
\begin{equation}
 \label{prob.death}
 \Pr( \mbox{death at node $\eta_i$} ) = \frac{D_{i}(T, \theta_T)}{ B(T, \theta_T) + D(T, \theta_T) }.
\end{equation} 

\noindent
The corresponding Markov process converges to the target posterior distribution in Equation \ref{eq:target posterior} given sufficient conditions that are provided in the following theorem.  
\begin{theorem}
\label{theorem:bd}
The birth-death process defined by the birth and death probabilities in Equations \ref{prob.birth} and \ref{prob.death} has stationary distribution $Pr(T,\theta_{T}\mid \mathcal{D})$, provided birth and death rates satisfy
\begin{eqnarray*}
B_{ijk}((T,\theta_{T}) Pr(T,\theta_{T}\mid \mathcal{D}) Pr(\mu^{n}_l) Pr( \mu^{n}_r ) = D_{i}(T^{b_{ijk}},\theta_{T^{b_{ijk}}}) Pr( \mu_{i} ) \Pr(T^{b_{ijk}},\theta_{T^{b_{ijk}}} \mid \mathcal{D}).
\end{eqnarray*} 
\end{theorem}
{\bf Proof.} Our proof draws on the theory of general continuous-time Markov birth-death processes derived by \cite[Section 7 and 8]{preston1976}. The process evolves by jumps which occur a finite number of times in any finite time interval. The jumps are of two types: a \textit{birth} in which the process jumps to a state with the additional point, whereas a \textit{death} in which the process jumps to a state with one less point by deleting one of the points in the current state. For the general case,  \citet[Theorem 7.1]{preston1976} proves the process converges to the target stationary distribution if the detailed balance conditions hold, as described in Theorem~\ref{theorem:bd}. For the case of a decision tree, if a birth occurs then we add one node to the current tree and if a death occurs we remove one node from the current tree. We design the CT-MCMC search algorithm in such a way that the stationary distribution of the process is our target posterior distribution Equation \ref{eq:target posterior}.  For a detailed proof see the Appendix~\ref{app:theorem}. 

Based on Theorem \ref{theorem:bd}, we can derive the birth and death rates as a function of the ratio of the target posterior distribution as follows
\begin{equation}
\label{eq:birth}
B_{ijk}( T, \theta_T ) = \min \left\{ 1, \frac{ \Pr(T^{b_{ijk}},\theta_{T^{b_{ijk}}} \mid \mathcal{D}) Pr( \mu_{i} ) }{ Pr(T,\theta_{T}\mid \mathcal{D}) Pr(\mu^{n}_l) Pr( \mu^{n}_r )  }  \right\}
\end{equation}
and 
\begin{equation}
\label{eq:death}
D_{i}( T, \theta_T ) = \min \left\{ 1, \frac{ \Pr(T^{d_{i}},\theta_{T^{d_{i}}} \mid \mathcal{D}) Pr(\mu^{i}_l) Pr( \mu^{i}_r ) }{ Pr(T,\theta_{T}\mid \mathcal{D})  Pr( \mu_{n} ) }  \right\}.
\end{equation}
Given the results provided above, our proposed algorithm for the posterior sampling from a (sums of) tree model is presented in the Algorithm~\ref{alg:CT-MCMC}.  

\begin{figure}[ht]
\includegraphics[width=1\textwidth]{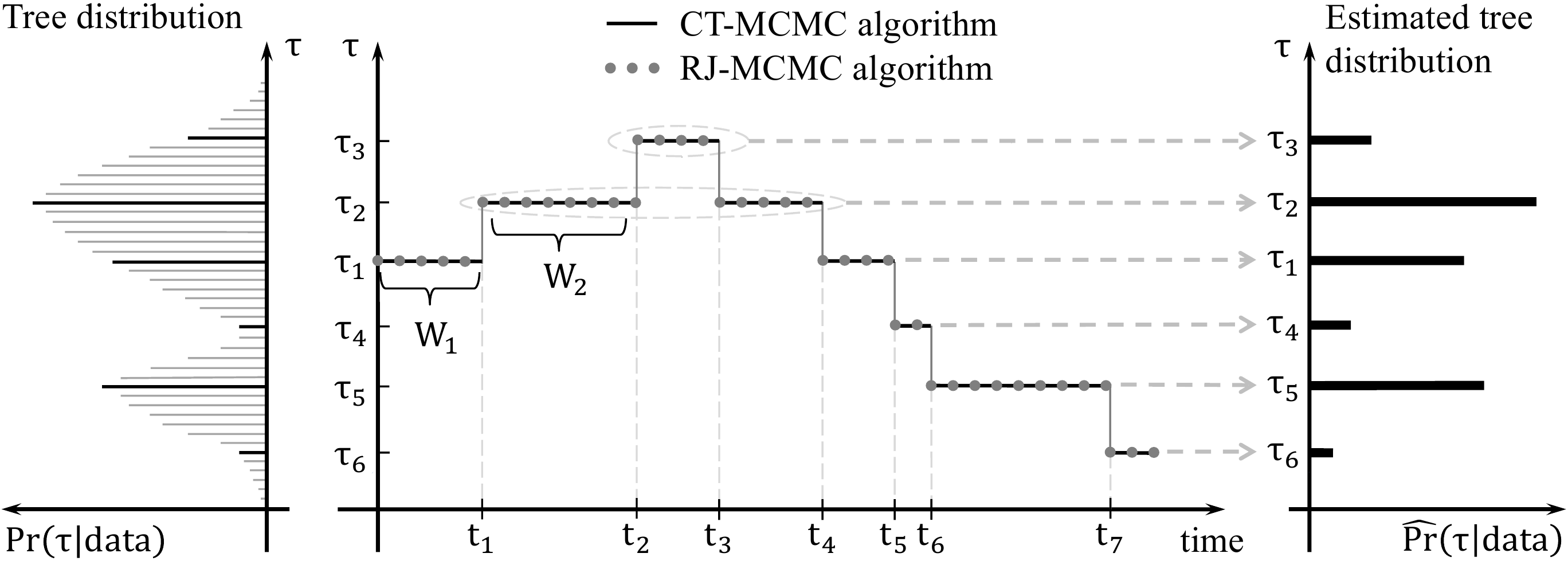}
\caption{Graphical representation of the CT-MCMC algorithm versus RJ-MCMC algorithm. The left panel presents the target posterior distribution of the tree topologies. The middle panel shows the sampling scheme of CT-MCMC and RJ-MCMC search algorithms. CT-MCMC algorithm samples in continuous time and $\left\{ W_1, W_2,... \right\}$ stand for waiting times (or holding times) and $\left\{ t_1, t_2,... \right\}$ stand for jumping times of the CT-MCMC algorithm; while the RJ-MCMC algorithm samples in discrete-time and dots visualize the sampling scheme of the algorithm. The right panel shows the estimated posterior distribution of the tree topologies based on the CT-MCMC sampler which are the proportional to the total waiting times of the visited trees, according to the Rao-Blackwellized estimator; see Subsection \ref{Posterior estimation}. }
 \label{fig:CT-MCMC} 
\end{figure}

\begin{algorithm}
\renewcommand{\algorithmicrequire}{\textbf{Input:}}
\renewcommand{\algorithmicensure}{\textbf{Output:}} 
\caption{. CT-MCMC search algorithm}
\label{alg:CT-MCMC}
\begin{algorithmic}[0]
\REQUIRE A tree $(T,\theta_T)$, data $\mathcal{D}$.
\FOR {$N$ iterations}
  \FOR {all the possible moves (for $i \in 1, ..., n_t$, $j \in 1, ..., n_{\nu}$, $k \in 1, ..., n_c$) in parallel}
   \STATE Draw the new split rules $(\nu_j, c_k)$.
   \STATE Draw the new $\mu_i$'s.
   \STATE Calculate the birth rates $B_{ijk}(T, \theta_T)$ and  death rates $D_i(T, \theta_T)$ according to Equations \ref{eq:birth}  and \ref{eq:death}.
  \ENDFOR
  \STATE Calculate the waiting time $W(T, \theta_T)$ given by Equation \ref{waiting time}.
  \STATE Update the new topology $\tau$ based on birth/death probabilities in Equations \ref{prob.birth} and \ref{prob.death}.
 \STATE Update $\sigma^2$ using standard Gibbs sampling scheme.
\ENDFOR
\ENSURE Samples from the full posterior distribution, Equation \ref{eq:full posterior}. 
\end{algorithmic}
\end{algorithm}
Algorithm~\ref{alg:CT-MCMC} presents the pseudo-code for the CT-MCMC search algorithm which samples from the posterior distribution in Equation \ref{eq:full posterior} by using the continuous-time birth-death mechanism that is described above. Basically, in the CT-MCMC search algorithm, we only simulate the jump chain and store each tree which visits and the corresponding waiting time. For the graphical visualization of the algorithm see Figure~\ref{fig:CT-MCMC}. 

One important feature of the CT-MCMC search algorithm is that a continuous time jump process is associated with the birth and death rates (Equations \ref{eq:birth} and \ref{eq:death}): whenever a jump occurs, the corresponding move is always accepted. In fact, the acceptance probability of usual RJ-MCMC search algorithm is replaced by the waiting times \eqref{waiting time} in the CT-MCMC search algorithm. Particularly, implausible trees, i.e. trees with low posterior probabilities have small waiting times and as a result die quickly; Conversely, plausible trees, i.e. trees with high posterior probabilities have larger waiting times. Thus, the CT-MCMC search algorithm are efficient to detect the high posterior probabilities regimes particularly for high-dimensional space models. 

\subsection{Computational improvements and further additions}

The key computational bottleneck of the CT-MCMC search algorithm is the computation of the birth and death rates over all the possible moves of the next step; The number of all possible moves exponentially increases with respect to the size of the tree-topology. Fortunately, in each step of the search algorithm, the birth and death rates can be calculated independently of each other; Thus, the rates can be computed in parallel which represents a key computational improvement of the CT-MCMC search algorithm with respect to RJ-MCMC. We implement this step of the above algorithm in parallel using using OpenMP in C to speed up the computations.

While Algorithm~\ref{alg:CT-MCMC} is feasible, in practice, it can be improved by
\begin{enumerate}[a)]
\item exploiting conjugacy,
\item including rotation proposals \citep[as initially suggested by][for the RJ-MCMC case]{pratola2016efficient}.
\end{enumerate}
 Below we detail each in turn.

Conjugate priors on the terminal node parameters $\mu_i\in \Theta_T, \ \ i \in \{1,\ldots,n_t\}$, can simplify the CT-MCMC algorithm. In the example below we are interested in modeling a continuous response which leads to i.i.d. priors $\pi(\mu_i)\sim N(0,\tau^2)$  \citep[][]{chipman2010bart}.  Marginalizing out a single terminal node parameter $\mu_i$, the integrated likelihood is given by 
\[
Pr(T,\overline{\theta_T}\vert\mathcal{D})=\int_{\mu_i}Pr(T,\theta_{T})\pi(\mu_i)d\mu_i
\]
which is available in closed form for conjugate priors (similarly for integrating two terminal node parameters).  Applying this marginalization to Equations \ref{eq:birth} and \ref{eq:death}, the updated birth and death rates for CT-MCMC search algorithm are
\begin{equation}
\label{eq:birth_marg}
B_{ijk}( T, \overline{\theta_T} ) = \min \left\{ 1, \frac{ \Pr(T^{b_{ijk}},\overline{\theta_{T^{b_{ijk}}}} \mid \mathcal{D})}{ Pr(T,\overline{\theta_{T}}\mid \mathcal{D}) }  \right\}
\end{equation}
and 
\begin{equation}
\label{eq:death_marg}
D_{i}( T, \overline{\theta_T} ) = \min \left\{ 1, \frac{ \Pr(T^{d_{i}},\overline{\theta_{T^{d_{i}}}} \mid \mathcal{D}) }{ Pr(T,\overline{\theta_{T}} \mid \mathcal{D})}  \right\}.
\end{equation}
Considering the above birth and death rates, we present our implemented CT-MCMC search algorithm in Algorithm \ref{alg:CT-MCMC-2}.
\begin{algorithm}
\renewcommand{\algorithmicrequire}{\textbf{Input:}}
\renewcommand{\algorithmicensure}{\textbf{Output:}} 
\caption{. CT-MCMC search algorithm - exploiting conjugacy}
\label{alg:CT-MCMC-2}
\begin{algorithmic}[0]
\REQUIRE A tree $(T, \overline{\theta_T})$, data $\mathcal{D}$.
\FOR {$N$ iterations}
  \FOR {all the possible moves (for $i \in 1, ..., n_t$, $j \in 1, ..., n_{\nu}$, $k \in 1, ..., n_c$) in parallel}
   \STATE Draw the new split rules $(\nu_j, c_k)$.
   \STATE Calculate the birth rates $B_{ijk}( T, \overline{\theta_T} ) $ and  death rates $D_i( T, \overline{\theta_T} ) $ according to Equations \ref{eq:birth_marg}  and \ref{eq:death_marg}.
  \ENDFOR
  \STATE Calculate the waiting time $W(T, \overline{\theta_T})$ given by Equation \ref{waiting time}, using Equations \ref{eq:birth_marg}  and \ref{eq:death_marg}.
  \STATE Update the new topology $\tau$ based on birth/death probabilities in Equations \ref{prob.birth} and \ref{prob.death}, using Equations \ref{eq:birth_marg}  and \ref{eq:death_marg}.
  \STATE Update $\sigma^2$ using standard Gibbs sampling scheme.
\ENDFOR
\ENSURE Samples from the full posterior distribution, Equation \ref{eq:full posterior}. 
\end{algorithmic}
\end{algorithm}

By integrating out the terminal node parameters $\mu_i\in \Theta_T, \ \ i \in \{1,\ldots,n_t\}$ in our tree model, we essentially exclude a sampling step inside the nested for loop in the Algorithm \ref{alg:CT-MCMC-2}; Thus, this algorithm is computationally more efficient than Algorithm~\ref{alg:CT-MCMC}.  

While until now we have introduced our main results focusing merely on birth-death moves for simplicity, building on recent work by \citet{pratola2016efficient} we can extend our sampling approach to so-called rotate proposals: rotate proposals can be thought of as a multivariate generalization of the simple univariate rotation mechanism found in the binary search tree literature \citep[see, e.g.,][]{sleator1988rotation} and implemented in \citet{gramacy2008bayesian}. This generalization allows dimension-changing proposals to occur at any interior node of a tree, and directly moves between modes of high likelihood and is described in detail in \citet{pratola2016efficient}. In Appendix~\ref{app:extention}, we demonstrate the correctness of this approach once added to the proposed birth-death mechanism in the CT-MCMC case. Moreover, we present an efficient way of implementing rotate proposals within Algorithm~\ref{alg:CT-MCMC} and \ref{alg:CT-MCMC-2} using marginalization.

\subsection{Posterior inference by samples in continuous time}
\label{Posterior estimation}

Figure~\ref{fig:CT-MCMC} shows the sampling scheme of CT-MCMC versus RJ-MCMC algorithm and how to estimate posterior quantities of interest using sampling in continuous time, based on model averaging.

Basically, for the case of discrete time RJ-MCMC sampler, we monitor its output after each iteration. In this case, based on model averaging, the posterior means are estimated by sample means
\begin{equation}
\label{eq:estimation}
E \left[ g(T, \theta_T) \right] \approx \dfrac{1}{N} \sum_{i=1}^{N} g( T_i, \theta_{T_i} )
\end{equation}
in which $N$ is the number of MCMC iterations. For the CT-MCMC sampler, at each jump, we store each state that it visits and the corresponding waiting time which are $\left\{ W_1, W_2,... \right\}$ in Figure~\ref{fig:CT-MCMC}. Note that alternative sampling schemes have been proposed -- for instance, similar to \cite{stephens2000bayesian}, the process may be sampled at regular times; see \cite{cappe2003reversible}.

We use the Rao-Blackwellized estimator \citep{cappe2003reversible} to estimate parameters of the models, based on model averaging. It is proportional to the expectation of length of the holding time in that tree which is estimated as the sum of the waiting times in that tree. In this case,  the posterior means are estimated by sample means
\begin{equation}
\label{eq:Rao-Blackwellized}
 E\left[g(T, \theta_T)\right] \approx \dfrac{ \sum_{i=1}^{N} W_i \left( T_i, \theta_{T_i} \right) g( T_i, \theta_{T_i} ) }{ \sum_{i=1}^{N} W_i \left( T_i, \theta_{T_i} \right) }.
\end{equation}
Effectively, the Rao-Blackwellized estimator depends on the waiting times \eqref{waiting time} of the visited trees by the CT-MCMC sampler. The waiting times are calculated based on all the possible birth and death moves from the current state $(T, \theta_T)$; Therefore, the waiting times essentially capture all the possible moves in each step . Therefore, by containing the waiting times in the Rao-Blackwellized estimator, all possible moves are incorporated into our estimation, not only those that are selected. Moreover, according to the Rao-Blackwell theorem, the variances of estimators built from the sampler output are decreased \citep{cappe2003reversible}.

Note that the Rao-Blackwellized estimator is based on model averaging, which has the advantage that it provides a coherent way of combining results over different models. As a result, the estimation of the parameter of interest is not based on only one single tree. In fact, the estimation is based on the all trees that are visited by the MCMC search algorithm. 

\section{Empirical evaluation of our sampling approach}
\label{sec:evaluation}

\begin{figure}[ht]
\centering
\includegraphics[width=1\textwidth]{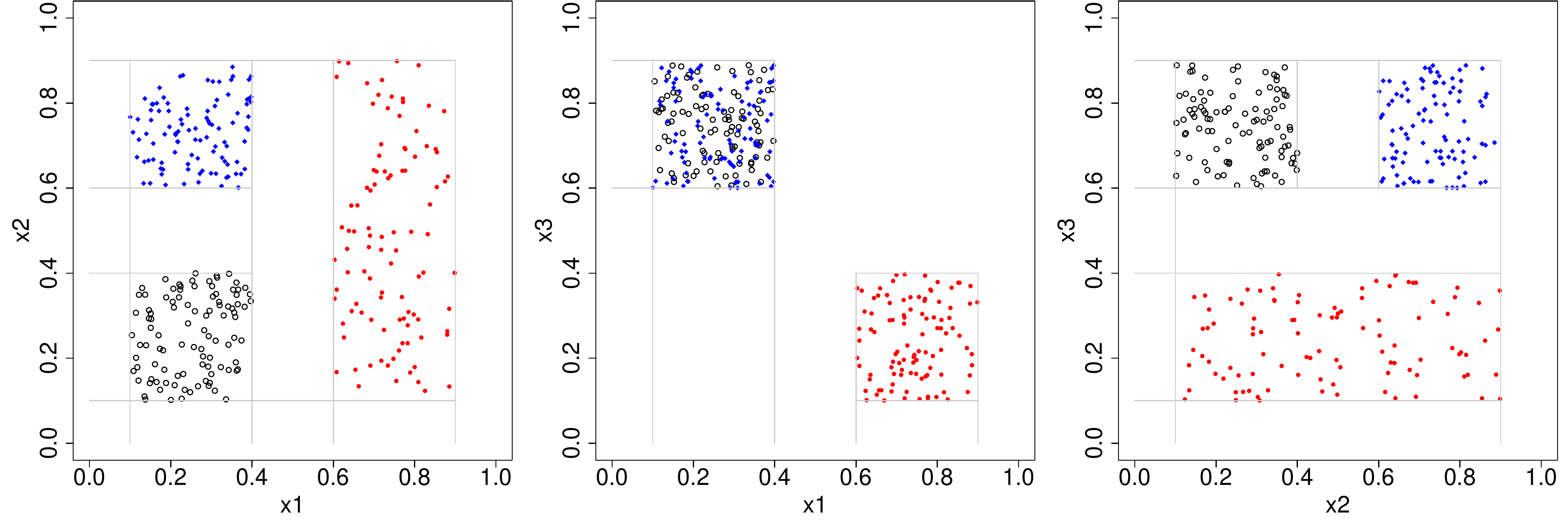}
\caption{Partition of the simulation data set with respect to the covariates $x_1$, $x_2$, and $x_3$ with the three regions defined in \ref{sim:x1}, \ref{sim:x2}, and \ref{sim:x3}. 
}
 \label{fig:sim-plot}
\end{figure}

We examine here the performance of the proposed CT-MCMC search algorithm based on a simulation scenario that is often used in the regression tree literature. This simulation scenario serves as a simple demonstration where proper mixing of the regression trees topological structure is important \citep{wu2007bayesian}. The synthetic data set consists of $n=300$ data points with $(x_1, x_2, x_3)$ covariates where
\begin{align}
\label{sim:x1}
x_{1i} \sim \begin{cases}
Unif(0.1, 0.4), \textrm{ for } i =  \{1, ..., 200\}   \\
Unif(0.6, 0.9), \textrm{ for } i =  \{201, ..., 300\}
\end{cases}
\end{align}

\begin{align}
\label{sim:x2}
x_{2i} \sim \begin{cases}
Unif(0.1, 0.4), \textrm{ for } i =  \{1, ..., 100\}\\
Unif(0.6, 0.9), \textrm{ for } i =  \{101, ..., 200\}\\
Unif(0.1, 0.9), \textrm{ for } i =  \{201, ..., 300\}
\end{cases}
\end{align}

\begin{align}
\label{sim:x3}
x_{3i} \sim \begin{cases}
Unif(0.6, 0.9), \textrm{ for } i =  \{1, ..., 200\}\\
Unif(0.1, 0.4), \textrm{ for } i = \{201, ..., 300\}
\end{cases}
\end{align}
Figure~\ref{fig:sim-plot} shows the partition of the simulation data set with respect to the covariates. Note that, following \citet{wu2007bayesian, pratola2016efficient}, we generate covariates such that the effects of $x_1$ and $x_3$ (see the middle panel in the Figure~\ref{fig:sim-plot}) are confounded which makes this data generating scheme particularly challenging. 

\begin{figure}[ht]
\centering
\includegraphics[width=1\textwidth]{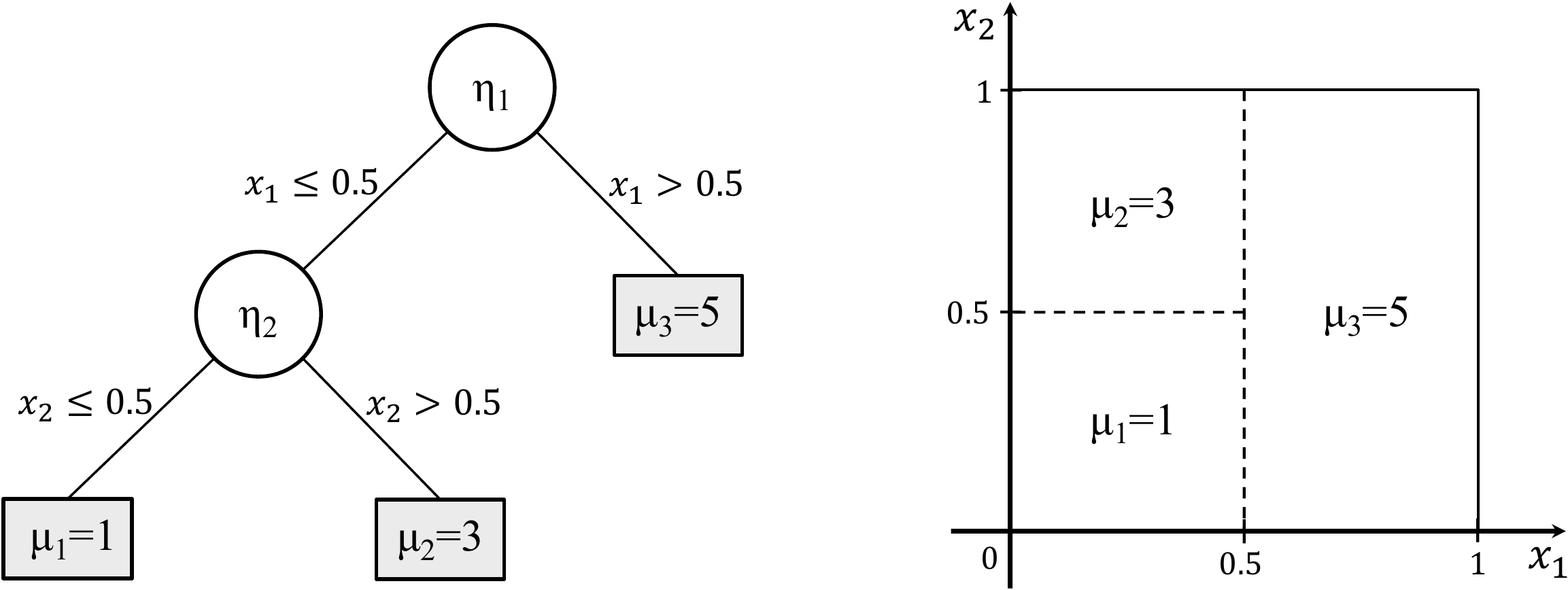}
\caption{True tree model (left) for the regression tree model in \ref{sim_data} where $y \sim \mathcal{N}(\mu,\sigma^2)$ with three partitions (right) based on the covariates $x_1$ and $x_2$.
}
 \label{fig:sim-tree}
\end{figure}

The response $y$ is calculated for $n = 300$ data points as:
\begin{align}
\label{sim_data}
y = \begin{cases}
1 + \mathcal{N}(0,\sigma^2) & \text{if } x_1 \leq 0.5, x_2 \leq 0.5\\
3 + \mathcal{N}(0,\sigma^2) & \text{if } x_1 \leq 0.5, x_2 > 0.5 \\
5 + \mathcal{N}(0,\sigma^2) & \text{if } x_1 > 0.5.
\end{cases}
\end{align}
Figure~\ref{fig:sim-tree} presents the above regression tree model with the partitions based on the covariates $x_1$ and $x_2$.

Following \citet{pratola2016efficient} we fit a single tree model (thus $M=1$ in Equation \ref{eq:sums}) to this data using the following approaches:
\begin{itemize}
\item \textbf{RJ-A}: Here we use a straightforward RJ-MCMC algorithm which is based on discrete time birth-death proposals as described in \citet{pratola2016efficient}. 
\item \textbf{RJ-B}: Here we use discrete time RJ-MCMC algorithm to which we add the rotation proposals as described in \citet{pratola2016efficient}.
\item \textbf{RJ-C}: Here we use discrete time RJ-MCMC algorithm including rotation proposals and perturbation. The latter addition concerns the second step of the sampling procedure as outlined in Section \ref{sec:notation} which concerns the sampling of the split rules $(\nu_i, c_i)$. This is not the main focus of this paper, however, we want to see whether this additional mechanism is also useful for the CT-MCMC approach  \citep[see][for detail]{pratola2016efficient}.
\item \textbf{CT-A}: Here we use our proposed CT-MCMC algorithm which is based on continuous-time birth-death approach but without perturbation; see Algorithm \ref{alg:CT-MCMC-2}.
\item \textbf{CT-B}: Here we add rotation proposals to the CT-MCMC algorithm described in  Algorithm \ref{alg:CT-MCMC-2}, again without perturbation; for details we refer to Appendix~\ref{app:extention}.
\item \textbf{CT-C}: Here we use both birth-death and rotation proposals and we add perturbation proposals to the second step of Algorithm \ref{alg:CT-MCMC-2}.
\end{itemize}

To evaluate the performance of the CT-MCMC search algorithm with compare with the RJ-MCMC, we run all the above search algorithms in the same conditions with 20,000 iterations and 1,000 as a burn-in. We perform all the computations in R and the computationally intensive tasks are implemented in parallel in C and interfaced in R. All the computations were carried out on an MacBook Pro with 2.9 GHz processor and Quad-Core Intel Core i7. 

For each of the above search algorithms, we report the following measurements: 
\begin{itemize}
\item \textbf{MSE}: This is the Mean Square Error. To calculate the MSE, we generate another synthetic data set consists of $n=300$ data points as a test set. Then, we compute the MSE of these test set based on the estimated tree models from the above MCMC search algorithms.
\item \textbf{Effective Sample Size}:  This is the number of effective independent draws that the algorithm generates. 
\item \textbf{Activity}: This is the proportion of splits on a given variable making up the tree decision rules.  In this synthetic example it is possible to derive the variable activity analytically which should be approximately $0.3$, $0.4$, and $0.3$ respectively. 
\item \textbf{Unique Trees}: The number of unique trees the algorithm generates.
\item \textbf{Effective Sample Size per Second}: The number of effective samples  drawn per second of computation time.
\end{itemize}

For the CT-MCMC approach, we estimate the parameter of interest based on the Rao-Blackwellized estimator in Equation \ref{eq:Rao-Blackwellized} and for the case of RJ-MCMC it is based on sample means in Equation \ref{eq:estimation}. Thus, all estimates are based on model averaging across all visited tree models; see Figure~\ref{fig:CT-MCMC}. 

Table \ref{tab:simresults1} presents the results for $\sigma^2 = 1$ which is a relatively challenging, high-noise, scenario. On average, over $100$ replications, the prediction error of each of the models is similar, and hence, as expected, the different sampling methods do not differentiate in terms of predictive performance. However, in terms of computational efficiency, measured as the effective sample size computed based on the posterior draws, it is clear that the CT-MCMC methods perform better than the RJ-MCMC approaches across the board. Our most elaborate proposal---combining CT-MCMC with both rotation proposals and perturbation proposals (CT-C) ---provides the best performance. This is especially prominent when looking at the exploration behavior of the different sampling methods as summarized by the number of unique trees visited. 
\begin{table}[ht!]
\centering
\begin{tabular}{c|c|cccc|HHHccc}
\toprule
\multicolumn{1}{c}{Method}& \multicolumn{1}{c}{ MSE}& \multicolumn{4}{c}{Effective Sample Size} 		& 		& 	   & 		    &  \multicolumn{3}{c}{Activity}  \\ 
		& 		& $\sigma^2$ & $x_1$ & $x_2$ & $x_3$ 	& system & time & user-time & $x_1$ & $x_2$ & $x_3$ 			\\	 
  \hline
 RJ-A & 1.02 & 19758 & 1037 & 2370 & 1265 & 0.0 & 0.75 & 0.75 & 0.27 & 0.45 & 0.28  \\ 
 RJ-B & 1.02 & 19774 & 1419 & 2899 & 1482 & 0.0 & 0.9 & 0.9 & 0.25 & 0.49 & 0.26  \\ 
 RJ-C & 1.02 & 19625 & 13134 & 1306 & 13144 & 0.0 & 1.72 & 1.72 & 0.25 & 0.5 & 0.25  \\ 
 CT-A & 1.02 & 19282 & 14160 & 32374 & 14128 & 0.01 & 3.59 & 3.58 & 0.28 & 0.43 & 0.28  \\ 
 CT-B & 1.02 & 19577 & 40041 & 74608 & 37925 & 0.01 & 3.55 & 3.54 & 0.27 & 0.48 & 0.25  \\ 
 CT-C & 1.02 & 20474 & 14452 & 51557 & 14518 & 0.02 & 4.44 & 4.41 & 0.26 & 0.47 & 0.26  \\ 
\bottomrule
\end{tabular}
\begin{tabular}{c|c|cccc}
\multicolumn{1}{c}{Method}& \multicolumn{1}{c}{Unique Trees}&  \multicolumn{4}{c}{Effective Sample Size per Second} \\
		&    & $\sigma^2$ & $x_1$ & $x_2$ & $x_3$ \\ 
  \hline
 RJ-A & 1.83 & 26344 & 1382 & 3160 & 1686 \\ 
 RJ-B & 3.07 & 21971 & 1576 & 3221 & 1646 \\ 
 RJ-C & 8.29 & 11410 & 7636 & 759 & 7642 \\ 
 CT-A & 11.44 & 5371 & 3944 & 9018 & 3935 \\ 
 CT-B & 3.82 & 5515 & 11279 & 21016 & 10683 \\ 
 CT-C & 11.71 & 4611 & 3255 & 11612 & 3270 \\ 
\bottomrule
\end{tabular}
\caption{Overview of the performance measures of different sampling methods for simulation example for the case $\sigma^2 = 1$ in Equation \ref{sim_data} . The table reports the average over $100$ replications of the prediction error, the sampling efficiency, the exploration behavior in terms of variable activity measured as the average proportion of internal rules involving each variable, the exploration behavior in terms of the average number of unique trees visited, and  computational efficiency in effective samples per second. }
\label{tab:simresults1}
\end{table}

Finally, it is clear that the computation time, measured as effective samples per second, of our newly proposed methods is on par, or faster, than the current state-of-art methods. To summarize, across the board we find a good empirical performance of our suggested CT-MCMC method(s). Appendix~\ref{app:simulation} provides additional simulation results for the cases $\sigma^2 \in (0.01, 0.1)$ showing that in both of these cases our suggested method again outperforms the RJ methods. In fact, in these lower noise scenarios the RJ methods fail to properly explore the parameters space while our suggested CT method maintains proper variable activity while better sampling tree-space.  We also note that the Rao-Blackwellization makes a greater impact in these low-noise scenarios, with an unweighted MSE of 0.014 for CT-C versus Rao-Blackwellized estimate of 0.01 reported in Table 2 and an unweighted MSE of 1.98E-3 for CT-C versus the Rao-Blackwellized estimate of 1.04E-4 reported in Table 3.

\section{Discussion}
\label{sec:discussion}

In this paper we introduced a continuous-time MCMC search algorithm for posterior sampling from Bayesian regression trees and sums of trees (BART). Our work is inspired by earlier work in this space demonstrating the efficiency of continuous-time MCMC search algorithms \citep[see, e.g.,][]{mohammadi2015bayesian, mohammadi2017bayesian}. Using the general theory described by \citet{preston1976} we have shown analytically that our proposed sampling approach converges to our desired target posterior distribution $\Pr(T,\theta_{T}\mid \mathcal{D})$ in the case of birth-death proposals. Next, we extended this result to also include the novel rotate proposals initially proposed by \citet{pratola2016efficient}. Jointly, these suggestions lead to an efficient sampling mechanism for Bayesian (additive) regression trees; a model that is gaining popularity in applied studies \citep[see, e.g.,][]{logan2019decision} and hence effective sampling methods are sought after.

The current work provides theoretical guarantees regarding the convergence of the CT-MCMC search algorithm. There is still room for additional computational improvements: while our marginalizing approach, combined with our mixture approach to include rotation proposals (see Appendix~\ref{app:extention}), provide important steps in providing a computationally feasible CT-MCMC method, we believe additional gains might be possible. Furthermore, while our current implementation parallelizes parts of the sampling process, additional gains might be achieved here. The current implementation of the methods proposed in this paper are available at \url{ https://bitbucket.org/mpratola/openbt}. 

We hope our current results improve the practical usability of Bayesian regression tree models for applied researchers by speeding up, and improving the accuracy, of the sampling process. Our methods seem to work well for reasonably sized problems (e.g., thousands of observations, tens of variables); we think their actual performance on big data sets needs to be further investigated.

\appendix

\section{Proof of Theorem \ref{theorem:bd}}
\label{app:theorem}

Our proof here is based on the theory of general continuous-time Markov birth-death processes derived by \cite{preston1976}.
We use the notation defined in the body of this paper. Assume that at a given time, the process is in a tree state $(T,\theta_T)$. The process is characterized by the {\it birth rates} $B_{ijk}(T, \theta_T)$, the {\it death rates} $D_{i}(T, \theta_T)$, and the birth and death \textit{transition kernels} $K_B((T, \theta_T) \to (T^*, \theta_{T^*}))$ and $K_D((T, \theta_T) \to (T^*, \theta_{T^*}))$. 

Birth and death events occur as independent Poisson processes with rates $B_{ijk}(T, \theta_T)$ and $D_{i}(T,\theta_T)$ respectively. Given that a specific birth occurs, the probability that the following jump leads to a point in $H \subset \Omega_{T^{b_{ijk}}}$ (where $\Omega_{T^{b_{ijk}}}$ is the space of $\theta_{T^{b_{ijk}}}$) is 
\begin{align*}
K_{B}((T, \theta_T) \to ( T^{b_{ijk}}, H )) &= Pr( T \to  T^{b_{ijk}}) \times Pr( \theta_{T^{b_{ijk}}} \to H |  T \to  T^{b_{ijk}} ) \nonumber \\
&= \frac{B_{ijk} (T, \theta_T)}{B(T, \theta_T)} \int \I( \theta_{T^{b_{ijk}}} \in H  )  Pr( \mu_{n_l} ) Pr( \mu_{n_r} ) d\mu_{n_l}  d\mu_{n_r}
\end{align*}
in which $B(T, \theta_T)=\sum_{ijk} B_{ijk}(T, \theta_T)$ and $Pr(.)$ is a proposal distribution for $\mu$'s.

Similarly, given a specific death occurs, the probability that the following jump leads to a point in $F \subset \Omega_{T^{d_{i}}}$ (where $\Omega_{T^{d_{i}}}$ is the space of $\theta_{T^{d_{i}}}$) is 
\begin{align}
 \label{kernel 2}
K_{D}((T, \theta_T) \to ( T^{d_{i}}, F )) &= Pr( T \to  T^{d_{i}}) \times Pr( \theta_{T^{d_{i}}} \to F | T \to  T^{d_{i}} ) \nonumber \\ 
&= \frac{D_{i} (T, \theta_T)}{D(T, \theta_T)} \int \I( \theta_{T^{d_{i}}} \in F  )  Pr( \mu_{i} ) d\mu_{i}
\end{align}
in which $D(T, \theta_T)=\sum_{i} D_{i}(T, \theta_T)$ and $Pr(.)$ is a proposal distribution for $\mu$'s.

This birth-death process satisfies the detailed balance conditions if
\begin{eqnarray}
\label{balance1}
\int_{H} B(T,\theta_T) \Pr(T, \theta_T\mid \mathcal{D}) d \theta_{T} = \\ \nonumber
\sum_{ijk} \int_{\theta_{T^{b_{ijk}}}} D(T^{b_{ijk}},\theta_{T^{b_{ijk}}}) K_{D}((T^{b_{ijk}}, \theta_{T^{b_{ijk}}}) \to ( T, F )) \Pr(T^{b_{ijk}},\theta_{T^{b_{ijk}}} \mid \mathcal{D}) d \theta_{T^{b_{ijk}}},
\end{eqnarray}
and
\begin{eqnarray*}
\int_{F} D(T,\theta_{T}) \Pr(T, \theta_{T} \mid \mathcal{D}) d \theta_{T} = \\  \nonumber
\sum_{i} \int_{\theta_{T^{d_{i}}}} B(T^{d_{i}}, \theta_{T^{d_{i}}}) K_{B}((T^{d_{i}}, \theta_{T^{d_{i}}}) \to ( T, F ))  \Pr(T^{d_{i}}, \theta_{T^{d_{i}}} \mid \mathcal{D}) d \theta_{T^{d_{i}}}.
\end{eqnarray*}

We check the first part of the detailed balance conditions (Equation \ref{balance1}) as follows. For the left hand side (LHS) we have
\begin{align*}
LHS &= \int_{F} B(T, \theta_{T}) \Pr(T,\theta_{T}\mid \mathcal{D}) d \theta_{T} \\
    &= \int_{\theta_{T}} \I(\theta_{T} \in F) B(T, \theta_{T}) \Pr(T,\theta_{T}\mid \mathcal{D}) d \theta_{T} \\
    &= \int_{\theta_{T}} \I(\theta_{T} \in F) \sum_{ijk}{B_{ijk}(T,\theta_{T})} \Pr(T,\theta_{T}\mid \mathcal{D}) d \theta_{T} \\
    &= \sum_{ijk} \int_{\theta_{T}} \I(\theta_{T} \in F) B_{ijk}(T,\theta_{T}) \Pr(T,\theta_{T}\mid \mathcal{D}) d \theta_{T} \\
    &= \sum_{ijk} \int_{\theta_{T}} \I(\theta_{T} \in F) B_{ijk}(T,\theta_{T}) \Pr(T,\theta_{T}\mid \mathcal{D}) \left[ \int_{(\mu^{n}_l,\mu^{n}_r)} Pr(\mu^{n}_l) Pr( \mu^{n}_r ) d\mu^{n}_l d\mu^{n}_r \right] d \theta_{T} \\
    & \qquad \qquad \qquad \qquad \qquad \qquad  \qquad \qquad \qquad \qquad \left[ Pr(.) \text{ must integrate to }1 \right] \\
&= \sum_{ijk} \int_{\theta_{T}} \int_{(\mu^{n}_l,\mu^{n}_r)}  \I(\theta_{T} \in F) B_{ijk}(T,\theta_{T}) \Pr(T,\theta_{T}\mid \mathcal{D}) Pr(\mu^{n}_l) Pr( \mu^{n}_r ) d\mu^{n}_l d\mu^{n}_r d \theta_{T}. \\    
\end{align*}
Furthermore, for the right hand side (RHS) of Equation \ref{balance1}, by using Equation \ref{kernel 2} we have
\begin{align*}
RHS &= \sum_{ijk} \int_{\theta_{T^{b_{ijk}}}} D(T^{b_{ijk}},\theta_{T^{b_{ijk}}}) K_{D}((T^{b_{ijk}}, \theta_{T^{b_{ijk}}}); ( T, F )) \Pr(T^{b_{ijk}},\theta_{T^{b_{ijk}}} \mid \mathcal{D}) d \theta_{T^{b_{ijk}}} \\    
    &  \qquad \qquad \qquad \qquad \qquad \qquad  \qquad \qquad \qquad \qquad \qquad \left[ \text{Equation \ref{kernel 2} } \right] \\
    & = \sum_{ijk} \int_{\theta_{T^{b_{ijk}}}} D_{i}(T^{b_{ijk}},\theta_{T^{b_{ijk}}}) \left[ \int_{\mu_i} \I( \theta_{T} \in F  )  Pr( \mu_{i} ) d\mu_{i} \right] \Pr(T^{b_{ijk}},\theta_{T^{b_{ijk}}} \mid \mathcal{D}) d \theta_{T^{b_{ijk}}} \\
    & = \sum_{ijk} \int_{\theta_{T^{b_{ijk}}}} \int_{\mu_i} \I( \theta_{T} \in F  )  D_{i}(T^{b_{ijk}},\theta_{T^{b_{ijk}}}) Pr( \mu_{i} ) \Pr(T^{b_{ijk}},\theta_{T^{b_{ijk}}} \mid \mathcal{D}) d\mu_{i}  d \theta_{T^{b_{ijk}}}. \\
\end{align*}
Note that the number of terminal nodes $n_t$ for performing a birth in the original tree $T$ equals the number of ways we can return by deaths $n_d.$

It follows that we have LHS=RHS provided that
\begin{eqnarray*}
B_{ijk}((T,\theta_{T}) Pr(T,\theta_{T}\mid \mathcal{D}) Pr(\mu^{n}_l) Pr( \mu^{n}_r ) = D_{i}(T^{b_{ijk}},\theta_{T^{b_{ijk}}}) Pr( \mu_{i} ) \Pr(T^{b_{ijk}},\theta_{T^{b_{ijk}}} \mid \mathcal{D}).
\end{eqnarray*} 

\section{Extending of CT-MCMC algorithm to rotate mechanism}
\label{app:extention}

Here we consider extending the CT-MCMC algorithm to include the rotate mechanism.  Following the construction of \citet{preston1976}, let the state space be $\Omega=\cup_{n=0}^\infty\Omega_{n}$ where $\Omega_{n}$ is made up of all states of cardinality $n$ and are disjoint. Further, let $\Omega^{b(n)}$ be the states from which a birth into $\Omega_{n}$ originates, let $\Omega^{d(n)}$ be the states from which a death into $\Omega_{n}$ originates and let $\Omega^{r(n)}$ be the states from which a rotate into $\Omega_{n}$ originates where $\Omega^{b(n)},\Omega^{d(n)},\Omega^{r(n)}$ are disjoint; that is $\Omega^{b(n)}\equiv\Omega_{n-1}$, $\Omega^{d(n)}\equiv\Omega_{n+1}$ and $\Omega^{r(n)}\subset\Omega\setminus ( \Omega^{b(n)}  \cup \Omega^{d(n)} ).$

Let $\mathcal{F}_n$ be the $\sigma$-field  of subsets of $\Omega_n$ and let $\mathcal{F}$ be the $\sigma$-field  on $\Omega$ generated by the $\mathcal{F}_n.$  We consider a jump process that can jump from state $x\in\Omega_n$ to a point in one of $\Omega^{b(n)},\Omega^{d(n)},\Omega^{r(n)}.$  Let $\mu$ denote a measure on $(\Omega,\mathcal{F})$ and $\mu_n$ denote $\mu$ restricted to $\Omega_n.$ Let $B,D,R:\Omega\rightarrow\mathbb{R}^+$ be $\mathcal{F}$-measurable with $D(x)=R(x)=0$ for $x\in\Omega_0$ and let $\alpha=B+D+R.$  For $n\geq 1$ we define the transition probability kernels
$$K_B^{(n)}:\Omega_n\times\mathcal{F}_{b(n)}\rightarrow\mathbb{R}^+,$$
$$K_D^{(n)}:\Omega_n\times\mathcal{F}_{d(n)}\rightarrow\mathbb{R}^+,$$
and
$$K_R^{(n)}:\Omega_n\times\mathcal{F}_{r(n)}\rightarrow\mathbb{R}^+.$$
Then the overall transition kernel is given by \citep{preston1976}
$$K(x,F)=\frac{B(x)}{\alpha(x)}K_B^{(n)}(x,F_{b(n)})+\frac{D(x)}{\alpha(x)}K_D^{(n)}(x,F_{d(n)})+\frac{R(X)}{\alpha(x)}K_R^{(n)}(x,F_{r(n)})$$
for $x\in\Omega_n, n\geq 1$ and let $\frac{B(x)}{\alpha(x)}=\frac{D(x)}{\alpha(x)}=\frac{R(X)}{\alpha(x)}=\frac{1}{2}$ if $\alpha(x)=0,$
and
$$K(x,F)=K_B^{(0)}(x,F_b(0))$$
if $x\in\Omega_0.$

A \emph{rotate event} goes to state $(T^{r_{ij}},  \theta_{T^{r_{ij}}})$ with \emph{rotate rate} $R_{ij}(T, \theta_T)$ where $i \in 1, ..., n_r$ and $n_r$ is the number of possible rotatable nodes \citep[see][for details]{pratola2016efficient} and $j \in 1,\ldots,n_j$ is the number of possible outcomes from a rotate at the  $i$'th rotatable node. Furthermore we define $R(T, \theta_T) =  \sum_{i=1}^{n_r}\sum_{j=1}^{n_i} R_{ij}(T, \theta_T)$. Hence, a rotate event changes the topology $\tau$ by rearranging internal nodes according to the rules described in \citet[][]{pratola2016efficient}.  

In total, we consider the overall number of topological changes to the tree to occur via birth and death moves (as defined earlier) and rotate moves which occur with respective rates $B_{ijk}(T,\theta_T), D_i(T,\theta_T)$ and $R_{ij}(T,\theta_T)$ given the tree is in state $(T,\theta_T)$. With rotate, we do not know how many of the $j$ possible outcomes of a rotate at node $i$ will increase the dimension of $\theta_T$ thereby creating a new $\mu$ parameter. So, to make things easier---and since this is what we do in practice---we integrate out all of these parameters and work directly with the marginal likelihood. In this case, the birth/death transition kernels from above become:
\begin{align*}
K_{B}(T \to T^{b_{ijk}}) &= \frac{B_{ijk} (T)}{B(T)}  , 
\end{align*}
\begin{align*}
K_{D}(T \to T^{d_{i}}) &= \frac{D_{i} (T)}{D(T)},  
\end{align*}
and
\begin{align*}
K_{R}(T \to T^{r_{ij}}) &= \frac{R_{ij} (T)}{R(T)}.  
\end{align*}

One of the things we need is that birth is inverse of death, death is inverse of birth and rotate is inverse of rotate. This means that in this case our detailed balance condition will consist of 3 equations, essentially the birth/death balances from earlier as well as a rotate balance condition
\begin{align*}
B(T) \Pr(T\mid \mathcal{D}) = \sum_{ijk} D(T^{b_{ijk}})K_D((T^{b_{ijk}})\to T)\Pr(T^{b_{ijk}}\mid\mathcal{D}),
\end{align*}
\begin{align*}
D(T) \Pr(T\mid \mathcal{D}) &= \sum_{i} B(T^{d_{i}})K_B((T^{d_{i}})\to T)\Pr(T^{d_{i}}\mid\mathcal{D}),
\end{align*}
and
\begin{align*}
R(T) \Pr(T\mid \mathcal{D}) &= \sum_{ij} R(T^{r_{ij}})K_R((T^{r_{ij}})\to T)\Pr(T^{r_{ij}}\mid\mathcal{D}),
\end{align*}
where $T^r_{ij}$ is the tree state generated from previously choosing the $j$'th rotate generated at rotatable node $i$ and $Pr(T\mid \mathcal{D})=\int_{\theta^T}Pr(T,\theta_T\mid \mathcal{D})$ is the marginal posterior.

For the rotate balance, we have
\begin{align*}
R(T) \Pr(T\mid \mathcal{D}) &= \sum_{ij} R(T^{r_{ij}})K_R(T^{r_{ij}}\to T)\Pr(T^{r_{ij}}\mid\mathcal{D})\\
\sum_{ij} R_{ij}(T) \Pr(T\mid \mathcal{D} ) &= \sum_{ij} R_{ij(T^{r_ij})} P(T^{r_ij}\mid \mathcal{D})\\
\end{align*}
which is satisfied if
\begin{align*}
R_{ij}(T) P(T\mid\mathcal{D}) &= R_{ij}(T^{r_ij})P(T^{r_{ij}}\mid\mathcal{D}).
\end{align*}

Thus, the corresponding rate for the rotate move is
\begin{equation*}
R_{ij}( T ) = \min \left\{ 1, \frac{ \Pr(T^{r_{ij}} \mid \mathcal{D}) }{ Pr(T\mid \mathcal{D}) }  \right\}
\end{equation*}
and similarly working with the integrated posterior, the corresponding rates for the birth/death moves become
\begin{equation*}
B_{ijk}( T ) = \min \left\{ 1, \frac{ \Pr(T^{b_{ijk}} \mid \mathcal{D}) }{ Pr(T\mid \mathcal{D}) }  \right\}
\end{equation*}
and
\begin{equation*}
D_{i}( T ) = \min \left\{ 1, \frac{ \Pr(T^{d_{i}} \mid \mathcal{D}) }{ Pr(T\mid \mathcal{D}) }  \right\}.
\end{equation*}

Given this construction, the probability of birth, death and rotate moves occur with probabilities given by
\begin{equation*}
 \Pr( \mbox{birth at node $\eta_i$ for variable $\nu_j$ and cut-point $c_k$} ) = \frac{B_{ijk}(T)}{ B(T) + D(T) + R(T)}, 
\end{equation*}
\begin{equation*}
 \Pr( \mbox{death at node $\eta_i$} ) = \frac{D_{i}(T)}{ B(T) + D(T) + R(T) }
\end{equation*} 
and
\begin{equation*}
 \Pr( \mbox{rotate $j$ at node $\eta_i$} ) = \frac{R_{ij}(T)}{ B(T) + D(T) + R(T) }.
\end{equation*} 

Note that in practice this approach is too expensive because we have to calculate $B(T)+D(T)+R(T)$ at each iteration. To address this problem we split this move into two moves: a birth/death part and a rotate part can be performed separately to reduce computational burden.To do so, we introduce parameter $\alpha$. The idea is that with probability $\alpha \in [0, 1]$ we perform a birth/death move via CT-MCMC, and with probability $1-\alpha$ we perform a rotate move via CT-MCMC.  That is, our move corresponds to the mixture distribution 
\[
\alpha \left[\frac{B_{ijk}(T)}{B(T)+D(T)} + \frac{D_i(T)}{B(T)+D(T)} \right]+(1-\alpha)\frac{R_{ij}(T)}{R(T)}
\]
 for some fixed, known $\alpha.$ Note that if 
 \[
 \alpha=\frac{B(T)+D(T)}{B(T)+D(T)+R(T)}
 \] 
 then this mixture distribution corresponds exactly to the distribution for the full CT-MCMC algorithm.

\section{Additional simulation results}
\label{app:simulation}

Here we present a number of additional simulation results for the simulation scenario in the Section \ref{sec:evaluation} and described in the main text for $\sigma^2 \in (0.1, 0.01)$. Tables \ref{tab:simresults2} and \ref{tab:simresults3} demonstrate that also in these cases our proposed CT-MCMC method performs well.

\begin{table}
\centering
\begin{tabular}{c|c|cccc|HHHccc}
\toprule
\multicolumn{1}{c}{Method}& \multicolumn{1}{c}{ MSE}& \multicolumn{4}{c}{Effective Sample Size} 		& 		& 	   & 		    &  \multicolumn{3}{c}{Activity}  \\ 
		& 		& $\sigma^2$ & $x_1$ & $x_2$ & $x_3$ 	& system & time & user-time & $x_1$ & $x_2$ & $x_3$ 			\\	 
  \hline
 RJ-A & 0.01 & 19735 & 1279 & 2275 & 996 & 0.0 & 0.74 & 0.74 & 0.29 & 0.45 & 0.26  \\ 
 RJ-B & 0.01 & 19707 & 1203 & 3021 & 1818 & 0.0 & 0.91 & 0.9 & 0.23 & 0.47 & 0.30  \\ 
 RJ-C & 0.01 & 19660 & 13247 & 2141 & 13255 & 0.0 & 1.68 & 1.68 & 0.25 & 0.5 & 0.25  \\ 
 CT-A & 0.01 & 24759 & 14276 & 39968 & 14333 & 0.01 & 3.59 & 3.58 & 0.28 & 0.44 & 0.28  \\ 
 CT-B & 0.01 & 19703 & 36275 & 75526 & 39250 & 0.01 & 3.53 & 3.52 & 0.25 & 0.48 & 0.27  \\ 
 CT-C & 0.01 & 29061 & 14470 & 55343 & 14467 & 0.01 & 4.41 & 4.40 & 0.26 & 0.47 & 0.26  \\ 
\bottomrule
\end{tabular}
\begin{tabular}{c|c|cccc}
\multicolumn{1}{c}{Method}& \multicolumn{1}{c}{Unique Trees}&  \multicolumn{4}{c}{Effective Sample Size per Second} \\
		&    & $\sigma^2$ & $x_1$ & $x_2$ & $x_3$ \\ 
  \hline
 RJ-A & 1.67 & 26669 & 1729 & 3074 & 1346 \\ 
 RJ-B & 2.68 & 21656 & 1322 & 3319 & 1997 \\ 
 RJ-C & 7.98 & 11702 & 7885 & 1274 & 7890 \\ 
 CT-A & 10.24 & 6897 & 3977 & 11133 & 3993 \\ 
 CT-B & 3.00 & 5582 & 10276 & 21395 & 11119 \\ 
 CT-C & 10.0 & 6590 & 3281 & 12550 & 3280 \\ 
\bottomrule
\end{tabular}
\caption{Overview of the performance measures of different sampling methods for simulation example for the case $\sigma^2 = 0.1$ in Equation \ref{sim_data} . The table reports the average over $100$ replications of the prediction error, the sampling efficiency, the exploration behavior in terms of variable activity measured as the average proportion of internal rules involving each variable, the exploration behavior in terms of the average number of unique trees visited, and  computational efficiency in effective samples per second.
}
\label{tab:simresults2}
\end{table}

\begin{table}
\centering
\begin{tabular}{c|c|cccc|HHHccc}
\toprule
\multicolumn{1}{c}{Method}& \multicolumn{1}{c}{ MSE}& \multicolumn{4}{c}{Effective Sample Size} 		& 		& 	   & 		    &  \multicolumn{3}{c}{Activity}  \\ 
		& 		& $\sigma^2$ & $x_1$ & $x_2$ & $x_3$ 	& system & time & user-time & $x_1$ & $x_2$ & $x_3$ 			\\	 
  \hline
 RJ-A & 1.02E-4 & 19739 & 1410 & 3029 & 1620 & 0.01 & 0.74 & 0.74 & 0.25 & 0.44 & 0.31  \\ 
 RJ-B & 1.02E-4 & 19742 & 2687 & 5290 & 2603 & 0.0 & 0.88 & 0.88 & 0.29 & 0.49 & 0.22  \\ 
 RJ-C & 1.02E-4 & 19719 & 13315 & 3602 & 13319 & 0.0 & 1.68 & 1.67 & 0.25 & 0.5 & 0.25  \\ 
 CT-A & 1.03E-4 & 15230 & 14095 & 21063 & 14131 & 0.01 & 3.55 & 3.54 & 0.28 & 0.43 & 0.28  \\ 
 CT-B & 1.02E-4 & 19713 & 34644 & 75182 & 40538 & 0.01 & 3.53 & 3.52 & 0.24 & 0.48 & 0.28  \\ 
 CT-C & 1.04E-4 & 19254 & 14308 & 43675 & 14283 & 0.01 & 4.51 & 4.50 & 0.27 & 0.46 & 0.27  \\ 
\bottomrule
\end{tabular}
\begin{tabular}{c|c|cccc}
\multicolumn{1}{c}{Method}& \multicolumn{1}{c}{Unique Trees}&  \multicolumn{4}{c}{Effective Sample Size per Second} \\
		&    & $\sigma^2$ & $x_1$ & $x_2$ & $x_3$ \\ 
  \hline
 RJ-A & 1.61 & 26675 & 1905 & 4094 & 2189 \\ 
 RJ-B & 2.80 & 22434 & 3053 & 6011 & 2958 \\ 
 RJ-C & 7.71 & 11738 & 7926 & 2144 & 7928 \\ 
 CT-A & 15.8 & 4290 & 3971 & 5933 & 3981 \\ 
 CT-B & 3.00 & 5585 & 9814 & 21298 & 11484 \\ 
 CT-C & 10.0 & 4269 & 3173 & 9684 & 3167 \\ 
\bottomrule
\end{tabular}
\caption{Overview of the performance measures of different sampling methods for simulation example for the case $\sigma^2 = 0.01$ in Equation \ref{sim_data} . The table reports the average over $100$ replications of the prediction error, the sampling efficiency, the exploration behavior in terms of variable activity measured as the average proportion of internal rules involving each variable, the exploration behavior in terms of the average number of unique trees visited, and  computational efficiency in effective samples per second.
}
\label{tab:simresults3}
\end{table}

\newpage
\vskip 0.2in
\bibliography{references}

\end{document}